\documentclass[cameraready]{Interspeech}
\usepackage{booktabs}
\usepackage{multirow}
\usepackage{float}
\usepackage{graphicx}
\usepackage{amssymb}
\usepackage{array}
\usepackage{makecell}
\usepackage{svg}
\usepackage[most]{tcolorbox}
\usepackage{inconsolata} 
\usepackage{caption}
\usepackage{hyperref}

\title{Speak or Stay Silent: Context-Aware Turn-Taking in Multi-Party Dialogue}

\author[affiliation={1}]{Kratika}{Bhagtani}
\author[affiliation={2}]{Mrinal}{Anand}
\author[affiliation={2}]{Yu Chen (Robert)}{Xu}
\author[affiliation={2}]{Amit}{Kumar Singh Yadav}
\date{}

\address{
    $^1$ School of Electrical and Computer Engineering, Purdue University, United States\\
    $^2$ Ishiki Labs Inc.
}

\email{kbhagtan@purdue.edu, mrinal@ishikilabs.ai, robert@ishikilabs.ai,  amit@ishikilabs.ai}


\usepackage{comment}

\begin{document}

\maketitle

\begin{abstract}
Existing voice AI assistants treat every detected pause as an invitation to speak. This works in dyadic dialogue, but in multi-party settings, where an AI assistant participates alongside multiple speakers, pauses are abundant and ambiguous. 
An assistant that speaks on every pause becomes disruptive rather than useful.
In this work, we formulate context-aware turn-taking: at every detected pause, given the full conversation context, our method decides whether the assistant should speak or stay silent. We introduce a benchmark of over 120K labeled conversations spanning three multi-party corpora.
Evaluating eight recent large language models, we find that they consistently fail at context-aware turn-taking under zero-shot prompting. We then propose a supervised fine-tuning approach with reasoning traces, improving balanced accuracy by up to 23 percentage points. Our findings suggest that context-aware turn-taking is not an emergent capability; it must be explicitly trained.

\end{abstract}

\section{Introduction}

\begin{figure}[H]
    \centering
    \includegraphics[width=\columnwidth]{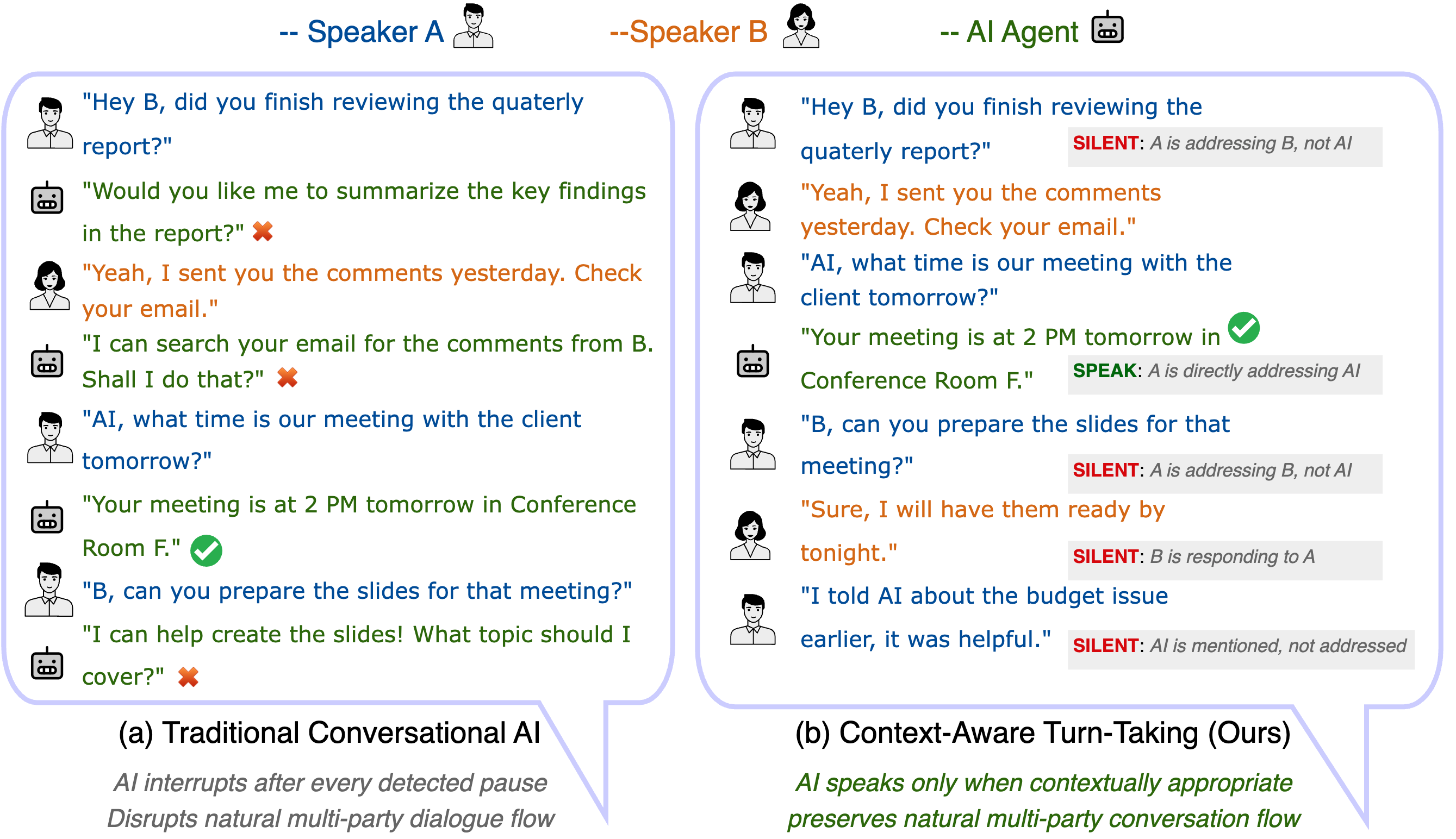}
    \caption{Traditional (a) vs. context-aware turn-taking (b) in Conversational AI. }
    \label{fig:figure-1}
\end{figure}

Large Language Models (LLMs) have rapidly advanced in instruction following, reasoning, and response generation~\cite{openai2024,qwen2025,penzo2024,li2023}, enabling their deployment as conversational Artificial Intelligence (AI) assistants~\cite{shuster2022,deng2023,zhang2020}. 
However, most dialogue systems and related evaluation benchmarks assume dyadic interactions between one user and one assistant~\cite{yi2025}. 
Real-world conversations are rarely dyadic. In meetings and group
conversations, an AI assistant participates alongside multiple speakers~\cite{wei2023,ouchi2016}.
In such settings the challenge shifts from \emph{what to say} to
\emph{whether and when to speak}~\cite{skantze2021,deng2023}. An AI assistant in a Zoom meeting that responds at every pause
becomes disruptive, while one that stays silent when addressed fails its role
(see Figure~\ref{fig:figure-1})~\cite{roddy2018,clark1991}.

Prior research on turn-taking has been largely focused on dyadic spoken dialogue - predicting turn boundaries from linguistic cues~\cite{ekstedt2020} or
identifying when an LLM should respond to a single user~\cite{umair2024} or when an audio LLM should proactively alert its user~\cite{yadav2026proactive}.
Speech LLMs~\cite{yadav2025speechnllama} take spoken input directly, and recent full-duplex speech models extend turn-taking to handle barge-in and backchannels, but both remain grounded in two-party, signal-level interaction~\cite{defossez2024,zhang2023speechgpt}.
Research in multi-party dialogue has been focused on structural sub-problems such as addressee recognition~\cite{le2019} and speaker-aware discourse parsing~\cite{yu2022}. 
None of these works address the integrated decision that a multi-party assistant must make at every pause, that is, given the full conversational context, should it speak or stay silent and not interrupt the conversation?


\vspace{-10pt}

In this work, we address this gap by formulating context-aware turn-taking as a supervised prediction task at every conversational pause. Our three major contributions are: 
(1) A benchmark containing more than 120K labeled decision points in conversations drawn from three multi-party corpora spanning workplace meetings~\cite{carletta2005,carletta2007}, social dialogue~\cite{wang2024friendsmmc}, and financial
conversations~\cite{grossman2025spgispeech20}. The decision points in the conversations are organized into four categories that capture explicit address, contextual intervention, and two forms of silence. 
(2) A large-scale evaluation of eight recent LLMs - including both closed-source~\cite{openai2024,team2024gemini,openai2025gpt52,google2024geminipro} and open-source~\cite{touvron2023llama,jiang2023mistral,qwen2025} models, and demonstrating that context-aware turn-taking fails under zero-shot prompting. 
(3) A supervised fine-tuning approach that uses distilled reasoning traces that improves balanced accuracy by up to 23 percentage points. 
\footnote{Code is available at \\
\href{https://github.com/ishikilabsinc/context_aware_modeling/tree/main}{https://github.com/ishikilabsinc/context\_aware\_modeling/tree/main}}

\section{Related Work}

Turn-taking has been widely studied in dyadic spoken dialogue.
Prior work predicts turn boundaries from text~\cite{ekstedt2020}
or near-future voice activity from audio using Voice Activity Projection (VAP)~\cite{ekstedt2022prosody}.
Recent studies also show that LLMs struggle with identifying Transition Relevance Places (TRPs) within a turn~\cite{umair2024}.
These approaches focus on signal-level turn shifts in two-party interaction.
In multi-speaker interaction, prior work has addressed sub-problems in isolation, such as speaker-aware discourse parsing~\cite{yu2022}, addressee recognition~\cite{le2019,inoue2025iwsds}, and response selection under multi-party structure~\cite{penzo2024}. 
Social intelligence benchmarks such as SocialEval~\cite{zhou2025socialeval} and AgentSense~\cite{leng2025agentsense} evaluate role consistency, goal completion, and interpersonal reasoning in multi-agent scenarios, but treat participation as a given rather than as a decision to be made at each conversational juncture.
Wei et al.~\cite{wei2023} highlight the importance of participation decisions in multi-party agents and introduce the MultiLIGHT dataset in a role-playing environment~\cite{wei2023}. 
However, because the dataset is derived from a fantasy role-playing game environment with assigned characters, it does not capture the dynamics of natural spoken conversations.
Moreover, the evaluation~\cite{wei2023} focuses on earlier conversational models such as BlenderBot~\cite{shuster2022}.
Hilgert and Niehues~\cite{hilgert2025} and MuPaS~\cite{wang2025mpft} address next-speaker prediction in multi-party dialogue.
Our work improves in three key respects. 
First, we formulate the task as a per-participant binary decision rather than predicting which speaker talks next, since an AI assistant deployed in a group conversation controls only its own participation.
Second, our benchmark is built from naturalistic multi-party corpora across three domains, with fine-grained labels distinguishing explicit and implicit contextual intervention or silence. Third, we evaluate recent LLMs to demonstrate that structured fine-tuning with added reasoning distillation yields substantial gains, a training paradigm not explored in prior work on this problem.

\section{Benchmark}
\label{sec:data}

In this section, we formulate the problem and benchmark.

\begin{table}[t!]
\centering
\small
\renewcommand{\arraystretch}{0.82}
\setlength{\tabcolsep}{6pt}
\begin{tabular}{lccccc}
\toprule
\textbf{Dataset} & \textbf{Total} & \textbf{I1} & \textbf{I2} & \textbf{S1} & \textbf{S2} \\
\midrule
AMI      & 11,900  & 1,598 & 4,407 & 4,127 & 1,768 \\
Friends  & 8,970   & 1,114 & 2,632 & 639 & 4,585 \\
SPGI     & 99,290  & 21,595 & 28,050 & 17,441 & 32,204 \\
\midrule
\textbf{Total} & \textbf{120,160} & \textbf{24,307} & \textbf{35,089} & \textbf{22,207} & \textbf{38,557} \\
\bottomrule
\end{tabular}
\caption{Distribution of datasets in our benchmark.}
\label{tab:dataset_distribution}
\vspace{-20pt}
\end{table}

\subsection{Problem Formulation}

We define context-aware turn-taking as follows. 
Given a multi-party conversation with \(N\) speakers, let \(C_t =(u_1, u_2, ..., u_t)\) denote the sequence of utterances up to time \(t\), where each utterance \(u_i\) is produced by some speaker \(s_i \in \{1, ..., N\}\).
After utterance \(u_t\), a pause is detected. 
For a designated target speaker \(k \neq s_t\), the goal is to predict a binary decision \(d_k \in \{\textsc{Speak}, \textsc{Silent}\}\) based on the conversational context \(C_t\).
This formulation is general in the sense that any participant can serve as the target speaker. 
During training, this allows us to derive supervision from naturally occurring human conversations since every speaker's behavior provides labeled examples. 
During inference, the target speaker is the AI assistant.

\subsection{Datasets and Benchmark Construction}
\label{sec:data_creation}
We construct the benchmark from three publicly available multi-party corpora spanning distinct conversational settings. These are described as follows.
\textbf{AMI Meeting Corpus}~\cite{carletta2005,carletta2007} contains approximately 100 hours of four-person design meetings with manual transcriptions and addressee annotations, covering questions and group discussions. 
We leverage the conversational annotations to infer addressee relationships and derive ground-truth for category assignment.
\textbf{Friends}~\cite{wang2024friendsmmc} provides transcripts from the television series, 
of scripted multi-party social dialogue typically involving 3–6 speakers.
\textbf{SPGISpeech 2.0}~\cite{grossman2025spgispeech20} contains transcribed earnings calls
and financial presentations with multiple participants. 
Together these datasets capture meeting-style interaction, informal social conversation, and domain-specific spoken dialogue.
\begin{figure}[!t]
    \centering
    \includegraphics[width=\linewidth]{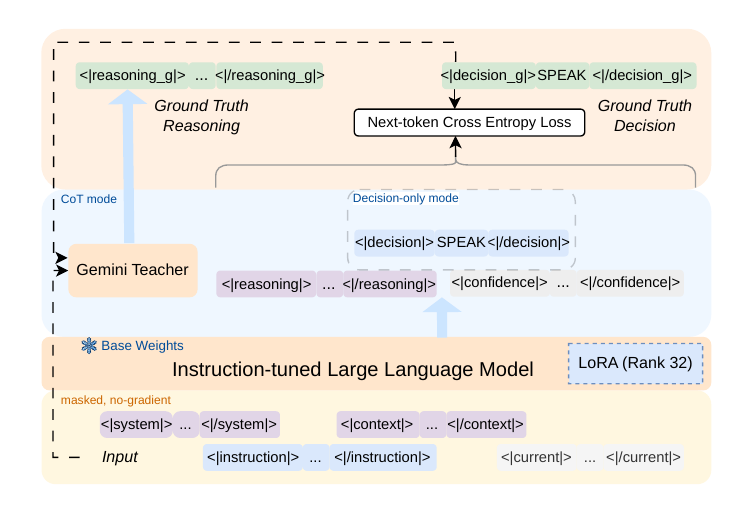}
    \caption{Overview of the proposed framework.}
    \label{fig:method}
    \vspace{-15pt}
\end{figure}
\begin{table*}[t]
\centering
\small
\renewcommand{\arraystretch}{0.82}
\setlength{\tabcolsep}{3pt}
\begin{tabular}{>{\raggedright\arraybackslash}p{1.6cm} l ccccccc}
\toprule
\multirow{2}{*}{\textbf{Model}} 
& \multirow{2}{*}{\textbf{Dataset}}
& \multicolumn{4}{c}{\textbf{Category-wise Accuracy (\%)}} 
& \multicolumn{3}{c}{\textbf{Overall (\%)}} \\
\cmidrule(lr){3-6} \cmidrule(lr){7-9}
& & \textbf{I1} & \textbf{I2} & \textbf{S1} & \textbf{S2}
& \textbf{Acc} & \textbf{F1$_{\text{avg}}$} & \textbf{Bal Acc} \\
\midrule

\multirow{3}{*}{\makecell[l]{gemini-3.1\\-pro}}
& AMI     
& 52.53 (-0.63) & 39.27 (-0.46) & 88.63 (+1.65) & \textbf{54.14} (+2.55)
& \textbf{60.77} (+0.68) & \textbf{59.55} (+0.52) & \textbf{61.03} (+0.70) \\

& Friends 
& 82.52 (-1.94) & 80.62 (-0.78) & 86.67 (+0.00) & \textbf{34.10} (+0.41)
& \textbf{56.43} (-0.22) & \textbf{56.11} (-0.17) & \textbf{60.54} (-0.36) \\

& SPGI    
& 79.35 (+0.14) & 30.31 (-1.58) & \textbf{91.32} (-0.17) & \textbf{69.07} (-0.13)
& \textbf{64.57} (-0.48) & \textbf{63.93} (-0.53) & \textbf{64.45} (-0.50) \\

\midrule

\multirow{3}{*}{gpt-5.2}
& AMI     
& 75.95 (-1.27) & 55.71 (-0.69) & 66.59 (-1.66) & 32.48 (-1.91)
& 59.23 (-0.93) & 59.20 (-0.95) & 59.21 (-0.95) \\

& Friends 
& 94.17 (+2.92) & 84.88 (+1.17) & 60.00 (+10.00) & 18.71 (-3.12)
& 49.00 (-0.33) & 46.63 (-0.76) & 55.41 (+0.00) \\

& SPGI    
& 85.30 (+0.00) & 31.54 (-0.90) & 75.48 (-0.71) & 49.45 (+0.28)
& 56.96 (-0.29) & 56.93 (-0.29) & 56.94 (-0.29) \\

\midrule

\multirow{3}{*}{\makecell[l]{gpt-oss\\-20b}}
& AMI     
& \textbf{\textit{74.05}} (+0.63) & \textbf{\textit{45.89}} (+3.43) & \textbf{\textit{65.40}} (+2.37) & \textbf{\textit{31.21}} (+5.10)
& \textbf{\textit{54.72}} (+2.90) & \textbf{\textit{54.72}} (+2.89) & \textbf{\textit{54.74}} (+2.90) \\

& Friends 
& 98.06 (-1.94) & 81.40 (+1.93) & 58.33 (-1.66) & 21.62 (-0.62)
& \textbf{\textit{49.89}} (-0.11) & \textbf{\textit{48.00}} (-0.28) & \textbf{\textit{55.92}} (+0.05) \\

& SPGI    
& 68.63 (+8.62) & 22.84 (+9.21) & 70.18 (-9.72) & 48.35 (-9.62)
& 49.62 (-0.46) & 49.35 (-0.21) & 49.54 (-0.35) \\

\midrule

\multirow{3}{*}{\makecell[l]{LLaMA3.1\\-8b-instruct}}
& AMI     
& \textbf{\textit{92.41}} (+1.89) & 69.86 (-1.14) & 30.57 (-2.61) & 9.55 (+0.64)
& 50.72 (-1.02) & 47.09 (-1.31) & 50.35 (-1.03) \\

& Friends 
& 97.09 (-2.92) & 85.27 (-0.39) & 50.00 (-1.67) & 16.01 (-2.29)
& 47.34 (-1.77) & 44.23 (-2.14) & 54.21 (-1.66) \\

& SPGI    
& \textbf{\textit{96.86}} (-0.98) & 65.32 (-3.54) & 20.75 (+1.59) & 4.58 (+1.59)
& 44.36 (-0.39) & 37.20 (+0.48) & 44.76 (-0.41) \\

\midrule

\multirow{3}{*}{\makecell[l]{Mistral-7b\\-instruct}}
& AMI     
& 89.24 (+1.27) & \textbf{\textit{83.33}} (+2.51) & 14.69 (+3.08) & 8.28 (+3.18)
& 49.45 (+2.64) & 41.59 (+3.24) & 48.93 (+2.64) \\

& Friends 
& 89.32 (+6.80) & 84.88 (+1.55) & 26.67 (+1.66) & 18.71 (+0.42)
& 46.23 (+1.55) & 43.30 (+1.40) & 52.87 (+1.80) \\

& SPGI    
& 84.60 (+3.23) & \textbf{\textit{72.03}} (+2.56) & 24.03 (-3.06) & 19.52 (-2.19)
& \textbf{\textit{49.01}} (+0.14) & \textbf{\textit{44.80}} (-0.78) & \textbf{\textit{49.33}} (+0.18) \\

\midrule

\multirow{3}{*}{\makecell[l]{Qwen2.5\\-7b}}
& AMI     
& 88.61 (-2.53) & 70.09 (-2.74) & 28.20 (+8.77) & 19.11 (+5.09)
& 50.72 (+2.47) & 47.34 (+3.87) & 50.37 (+2.54) \\

& Friends 
& 92.23 (+3.89) & 77.52 (-4.26) & 55.00 (+0.00) & 24.95 (+1.04)
& 49.67 (-0.22) & 48.39 (-0.02) & 55.00 (-0.51) \\

& SPGI    
& \textbf{\textit{89.28}} (-1.45) & 67.95 (+0.69) & 25.72 (+0.06) & 16.30 (+0.34)
& 48.15 (+0.00) & 43.68 (+0.08) & 48.48 (+0.00) \\

\midrule

\multirow{3}{*}{\makecell[l]{Qwen3\\-4b-instruct}}
& AMI     
& \textbf{\textit{92.41}} (+1.26) & 77.63 (+0.68) & 39.81 (+2.84) & 7.64 (-0.63)
& \textbf{\textit{56.68}} (+1.36) & \textbf{\textit{53.53}} (+1.59) & \textbf{\textit{56.32}} (+1.37) \\

& Friends 
& 100.00 (+0.00) & 90.70 (-0.78) & 38.33 (+0.00) & 6.03 (+1.45)
& 43.13 (+0.55) & 36.82 (+1.04) & 51.48 (+0.37) \\

& SPGI    
& 94.94 (-0.09) & 55.07 (-3.18) & 16.99 (-0.33) & 10.34 (-0.37)
& 42.25 (-1.09) & 36.81 (-0.89) & 42.60 (-1.10) \\

\midrule

\multirow{3}{*}{\makecell[l]{Qwen3\\-8b}}
& AMI     
& 90.51 (+0.63) & 68.95 (+0.68) & 48.82 (+0.00) & 6.37 (-0.64)
& 56.26 (+0.25) & 54.53 (+0.18) & 55.99 (+0.24) \\

& Friends 
& 100.00 (+0.00) & 87.21 (+1.55) & 43.33 (-3.33) & 6.44 (+1.25)
& 42.68 (+0.89) & 37.00 (+0.97) & 50.70 (+0.92) \\

& SPGI    
& 94.94 (-0.93) & 43.70 (-3.06) & 33.70 (-0.44) & 11.28 (+1.35)
& 42.46 (-0.70) & 39.31 (-0.34) & 42.73 (-0.72) \\

\bottomrule
\end{tabular}
\caption{
Zero-shot performance of evaluated LLMs on the context-aware turn-taking benchmark. Values in parentheses denote the change when the system prompt is repeated twice in the input; the base value uses a single system prompt. Bold indicates the best overall model, while bold+italic indicates the best open-source model.
}
\label{tab:baseline_comparison}
\vspace{-15pt}
\end{table*}
For each corpus, we segment transcripts into time-ordered sequences and generate one decision point per non-speaking participant at each utterance boundary. The ground-truth label is derived directly from the transcript: if speaker \(k\) produced the next utterance, the label is \textsc{Speak}; otherwise, \textsc{Silent}.
While the prediction task is binary, we assign each decision point a fine-grained category. The four categories capture qualitatively distinct situations:\\
\textbf{Explicit Address (I1):} The target speaker is directly
addressed by name or role, and is unambiguously expected to respond, making these the easiest cases for speaking (\textsc{Speak}).\\
\textbf{Contextual Intervention (I2):} The target is not referenced
but is an active participant and a response is expected (\textsc{Speak}). \\
\textbf{No Reference (S1):} The ongoing exchange involves other
speakers and the target remains a bystander (\textsc{Silent}).\\
\textbf{Referenced but not addressed (S2):} The target is
mentioned (e.g., in third person) but is not expected to respond (\textsc{Silent}). This category captures an important distinction that being talked \emph{about} is not the same as being talked \emph{to}.

\begin{table*}
\centering
\small
\renewcommand{\arraystretch}{0.82}
\setlength{\tabcolsep}{3pt}
\begin{tabular}{>{\raggedright\arraybackslash}p{1.4cm} l ccccccc}
\toprule
\multirow{2}{*}{\textbf{Model}}
& \multirow{2}{*}{\textbf{Dataset}}
& \multicolumn{4}{c}{\textbf{Category-wise Accuracy (\%)}} 
& \multicolumn{3}{c}{\textbf{Overall (\%)}} \\
\cmidrule(lr){3-6} \cmidrule(lr){7-9}
& & \textbf{I1} & \textbf{I2} & \textbf{S1} & \textbf{S2}
& \textbf{Acc} & \textbf{F1$_{\text{avg}}$} & \textbf{Bal Acc} \\
\midrule

\multirow{3}{*}{\makecell[l]{gpt-oss\\-20b}}
& AMI
& 70.89
& 50.00
& 62.56
& 22.93 ($\downarrow$ 12.69)
& 53.70
& 58.72
& 58.74 \\

& Friends 
& 93.20
& 84.11
& 65.00
& 20.37
& 49.89
& 49.10
& 56.84 \\

& SPGI    
& 76.50 
& 26.34
& 56.80 ($\downarrow$ 13.38)
& 29.43 ($\downarrow$ 18.93)
& 43.74 ($\downarrow$ 11.85)
& 49.63 
& 49.66 \\

\midrule

\multirow{3}{*}{\makecell[l]{Mistral-7b\\-instruct}}
& AMI
& 81.01
& 58.90 ($\downarrow$ 24.43)
& 86.49 (\textbf{$\uparrow$ 71.80})
& 61.78 (\textbf{$\uparrow$ 53.50})
& 72.17 (\textbf{$\uparrow$ 22.72})
& 72.05 (\textbf{$\uparrow$ 30.47})
& 72.28 (\textbf{$\uparrow$ 23.35}) \\

& Friends
& 98.06
& 52.33 ($\downarrow$ 32.56)
& 76.67 (\textbf{$\uparrow$ 50.00})
& 77.75 (\textbf{$\uparrow$ 59.04})
& 72.73 (\textbf{$\uparrow$ 26.50})
& 71.54 (\textbf{$\uparrow$ 28.24})
& 71.50 (\textbf{$\uparrow$ 18.63}) \\

& SPGI
& 59.22 ($\downarrow$ 25.38)
& 47.28 ($\downarrow$ 24.75)
& 71.27 (\textbf{$\uparrow$ 47.24})
& 60.11 (\textbf{$\uparrow$ 40.59})
& 58.39
& 58.22 (\textbf{$\uparrow$ 13.42})
& 58.33 \\

\midrule

\multirow{3}{*}{\makecell[l]{LLaMA3.1\\-8b-instruct}}

& AMI
& 80.38 ($\downarrow$ 12.03)
& 63.24
& 80.81 (\textbf{$\uparrow$ 50.24})
& 63.69 (\textbf{$\uparrow$ 54.14})
& 71.91 (\textbf{$\uparrow$ 21.19})
& 71.89 (\textbf{$\uparrow$ 24.80})
& 71.98 (\textbf{$\uparrow$ 21.62}) \\

& Friends
& 100.00
& 63.18 ($\downarrow$ 22.09)
& 86.67 (\textbf{$\uparrow$ 36.67})
& 69.44 (\textbf{$\uparrow$ 53.43})
& 72.28 (\textbf{$\uparrow$ 24.94})
& 71.78 (\textbf{$\uparrow$ 27.56})
& 72.52 (\textbf{$\uparrow$ 18.31}) \\

& SPGI
& 91.90
& 73.04
& 70.50 (\textbf{$\uparrow$ 49.75})
& 38.17 (\textbf{$\uparrow$ 33.59})
& 65.42 (\textbf{$\uparrow$ 20.85})
& 64.64 (\textbf{$\uparrow$ 27.44})
& 65.61 (\textbf{$\uparrow$ 20.85}) \\

\midrule

\multirow{3}{*}{\makecell[l]{Qwen2.5\\-7b}}

& AMI
& 91.14
& 68.72
& 74.64 (\textbf{$\uparrow$ 46.45})
& 52.87 (\textbf{$\uparrow$ 33.76})
& 71.74 (\textbf{$\uparrow$ 21.02})
& 71.70 (\textbf{$\uparrow$ 24.36})
& 71.70 (\textbf{$\uparrow$ 21.33}) \\

& Friends
& 100.00
& 74.03
& 88.33 (\textbf{$\uparrow$ 33.33})
& 47.19 (\textbf{$\uparrow$ 22.25})
& 63.64 (\textbf{$\uparrow$ 13.97})
& 63.63 (\textbf{$\uparrow$ 15.24})
& 66.60 (\textbf{$\uparrow$ 11.60}) \\

& SPGI    
& 97.10
& 81.16 (\textbf{$\uparrow$ 13.21})
& 62.86 (\textbf{$\uparrow$ 37.14})
& 32.50 (\textbf{$\uparrow$ 16.20})
& 65.58 (\textbf{$\uparrow$ 17.43})
& 63.91 (\textbf{$\uparrow$ 20.23})
& 65.83 (\textbf{$\uparrow$ 17.35}) \\

\midrule

\multirow{3}{*}{\makecell[l]{Qwen3-4b\\-instruct}}

& AMI
& 84.18
& 63.93 ($\downarrow$ 13.70)
& 79.62 (\textbf{$\uparrow$ 39.81})
& 60.51 (\textbf{$\uparrow$ 52.87})
& 71.83 (\textbf{$\uparrow$ 15.15})
& 71.82 (\textbf{$\uparrow$ 18.29})
& 71.87 (\textbf{$\uparrow$ 15.55}) \\

& Friends
& 100.00
& \textbf{57.75 ($\downarrow$ 32.95)}
& 96.67 (\textbf{$\uparrow$ 58.33})
& 55.93 (\textbf{$\uparrow$ 49.90})
& 64.19 (\textbf{$\uparrow$ 21.06})
& 63.94 (\textbf{$\uparrow$ 27.12})
& 65.12 (\textbf{$\uparrow$ 13.64}) \\

& SPGI
& 80.10 (\textbf{$\uparrow$ 17.85})
& 69.54 (\textbf{$\uparrow$ 14.47})
& 69.63 (\textbf{$\uparrow$ 39.82})
& 61.45 (\textbf{$\uparrow$ 52.87})
& 69.23 (\textbf{$\uparrow$ 26.98})
& 69.18 (\textbf{$\uparrow$ 32.37})
& 69.29 (\textbf{$\uparrow$ 26.69}) \\

\midrule

\multirow{3}{*}{\makecell[l]{Qwen3-8b\\-instruct}}

& AMI
& 74.68 ($\downarrow$ 15.82)
& 58.90 ($\downarrow$ 10.05)
& 83.89 (\textbf{$\uparrow$ 35.07})
& 69.43 (\textbf{$\uparrow$ 63.06})
& 71.40 (\textbf{$\uparrow$ 15.15})
& 71.25 (\textbf{$\uparrow$ 16.72})
& 71.53 (\textbf{$\uparrow$ 15.54}) \\

& Friends
& 100.00
& 47.67 ($\downarrow$ 39.53)
& 91.67 (\textbf{$\uparrow$ 48.33})
& 74.01 (\textbf{$\uparrow$ 67.57})
& 70.62 (\textbf{$\uparrow$ 27.94})
& 69.33 (\textbf{$\uparrow$ 32.33})
& 69.29 (\textbf{$\uparrow$ 18.59}) \\

& SPGI    
& 73.31 ($\downarrow$ 21.63)
& 64.89 (\textbf{$\uparrow$ 21.19})
& 57.89 (\textbf{$\uparrow$ 24.19})
& 48.39 (\textbf{$\uparrow$ 37.11})
& 60.11 (\textbf{$\uparrow$ 17.65})
& 59.87 (\textbf{$\uparrow$ 20.56}) 
& 60.20 (\textbf{$\uparrow$ 17.47}) \\

\bottomrule
\end{tabular}
\caption{Performance after supervised fine-tuning. 
Numbers in parentheses indicate the absolute change in percentage points relative to the zero-shot baseline in Table 2. 
Arrows are shown when the change exceeds ±10 points. 
Bold indicates improved performance.}
\label{tab:ft_comparison}
\vspace{-15pt}
\end{table*}

We remove filler-only utterances (e.g., ``um", ``uh-huh") and very short turns (fewer than 3 characters after removing punctuation), apply exact de-duplication to remove duplicate contexts.
All datasets are split into train/validation/test with an 80/10/10 ratio per-category. 
Samples with no context turns are filtered out.
For SPGISpeech, which is substantially larger than the other corpora, we apply stratified subsampling to approximately 11K training samples while preserving class and category proportions.
Table~\ref{tab:dataset_distribution} shows the distribution for the datasets after processing.
Examples from the dataset are presented in Appendix~\ref{app:dataset}. 
\footnote{Dataset is available at \\
\href{https://huggingface.co/datasets/ishiki-labs/multi-party-dialogue}{https://huggingface.co/datasets/ishiki-labs/multi-party-dialogue}}



\section{Proposed Method}\label{sec:proposed-method}

This section describes the proposed framework.

\subsection{Zero-Shot Prompting}\label{sec:benchmark}

We evaluate LLMs under zero-shot prompting.
Each model receives a system prompt describing the task.
We evaluate two closed-source models (gpt-5.2~\cite{openai2024,openai2025gpt52} and gemini~3.1-pro~\cite{team2024gemini,google2024geminipro}) and six open-source ones (gpt-oss-20b~\cite{openai2024}, LLaMA3.1-8b-instruct~\cite{touvron2023llama}, Mistral-7b-instruct~\cite{jiang2023mistral}, Qwen2.5-7b~\cite{qwen2025}, Qwen3-4b-instruct~\cite{qwen2025}, and Qwen3-8b~\cite{qwen2025}) at temperature 0. All models receive identical prompts and are evaluated on the test splits.
We evaluate model performance using both category-wise and aggregate metrics. Each decision point corresponds to a binary prediction
$\hat{d}_k \in \{\textsc{Speak}, \textsc{Silent}\}$ for the target
speaker $k$, where the ground-truth decision is $d_k$.
Let $D$ denote the set of decision points in the dataset.
We report three metrics:\\ (1) Accuracy ($\mathrm{Acc})=\frac{1}{|D|}\sum_{i\in D}\mathbf{1}[\hat{d}_i=d_i]$\\
(2) Class-averaged F1 ($\text{F1}_{\text{avg}})=\tfrac{1}{2}(\text{F1}_{\textsc{Speak}}+\text{F1}_{\textsc{Silent}})$, where $\text{F1}_c=\frac{2P_cR_c}{P_c+R_c}$ with precision $P_c$ and recall $R_c$ for class $c$.\\
(3) Balanced Accuracy
($\mathrm{BalAcc})=\tfrac{1}{2}(\mathrm{TPR}_{\textsc{Speak}}+\mathrm{TPR}_{\textsc{Silent}})$, where $\mathrm{TPR}_c$ denotes the True Positive Rate
(recall) for class $c$.
For diagnostic analysis,
we additionally report accuracy separately for the four benchmark
categories.

\subsection{Supervised Fine-Tuning (SFT)}
\label{sec:train_details}


We further propose supervised fine-tuning for the task. 
We fine-tune all open-source models using Low-Rank Adaptation (LoRA)~\cite{hu2022lora} (rank\( = \)32, \(\alpha=64\), dropout\( = \)0.05) on attention and Multilayer Perceptron (MLP) projection layers. Training uses AdamW optimizer with a learning rate \(10^{-4}\), cosine schedule, batch size 32 (16 \(\times\)2 steps of gradient accumulation), and 16-bit floating point precision for 3 epochs with 10 warmup steps, selecting the checkpoint with the best validation $\text{F1}_{\text{avg}}$.
During training, inputs are truncated to a maximum context length of 2048 tokens (most examples fit within this cap) for all models due to memory constraints except for gpt-oss-20b, which uses a limit of 1536 tokens due to the larger size and computational resource requirements of the model. 
When sequences exceed the limit, the most recent turns are retained. Fine-tuning uses 1–8 A100 80GB GPUs depending on model size (training completes in a few hours per dataset), with FSDP~\cite{zhao2023fsdp} for larger runs.
We train in two modes. In \textbf{Decision-only} mode, the model outputs only the binary decision. In \textbf{Reasoning with Decision} mode, the model first generates a one-sentence reasoning trace before the decision explaining why the target speaker should \(\textsc{Speak}\) or stay \(\textsc{Silent}\). 
To obtain reasoning traces for training, we use label-conditioned distillation: a teacher model (Gemini~2.5~Flash)~\cite{google2025gemini25flash} receives each training sample (system prompt, instruction prompt, context history and current turn) along with its ground-truth label and generates a one-sentence justification.
This conditioning ensures reasoning traces are consistent with the correct label while grounding explanations in observable dialogue context.
To prevent class and category imbalance from dominating training, we use a four-way balanced batch sampler that draws 25\% of each batch from each of the four categories (I1, I2, S1, S2). Figure~\ref{fig:method} shows an overview of our proposed framework.
The prompts used in Section~\ref{sec:proposed-method} are provided in the Appendix~\ref{app:prompts}. 
\footnote{Model checkpoints are available at \\
\href{https://huggingface.co/ishiki-labs/models}{https://huggingface.co/ishiki-labs/models}}




\section{Experimental Results}

\textbf{Experiment-1 Zero-shot Prompting \& SFT:}
Table~\ref{tab:baseline_comparison} reports zero-shot performance and shows that all models struggle. 
The best-performing model, gemini-3.1-pro, achieves only 64.45\% balanced accuracy on SPGI, while open-source models hover near random performance on all three datasets.
Most models exhibit a strong \textsc{Speak} bias, producing unacceptably low accuracies for S1 and S2. 
This confirms that context-aware turn-taking is not an inherent capability of instruction-tuned LLMs.
Repeating the system prompt twice~\cite{leviathan2025} yields marginal gains (\(\leq\)3 points, Table~\ref{tab:baseline_comparison}), confirming the failure reflects a fundamental lack of turn-taking capability, not instruction neglect.


Table~\ref{tab:ft_comparison} reports results after supervised fine-tuning with Decision-only mode.
Except for gpt-oss-20b, SFT yields substantial improvements across all models and datasets, with balanced accuracy gains of up to 23 percentage points. 
Mistral-7B-Instruct improves from 41.59\% F1$_{\text{avg}}$ in baseline to 72.05\% on AMI.
Gpt-oss-20b, a reasoning-oriented model, shows minimal gains from SFT. We attribute this to a conflict between its internalized chain-of-thought behavior and the LoRA-adapted output format. 
In effect, the adapter learns to produce the target format but cannot redirect the model's internal reasoning toward the task-specific pragmatic cues that drive gains in the other models.
Per-category analysis reveals that the largest gains come from S2 and S1, the two categories that require pragmatic reasoning to remain \textsc{Silent}. Models maintain performance on I1 (explicit address), which was already closer to near-perfect performance in Table~\ref{tab:baseline_comparison}.

\textbf{Experiment-2 Human Evaluation:}
\begin{table}[b!]
\vspace{-10pt}
\centering
\small
\resizebox{\linewidth}{!}{%
\begin{tabular}{l|cccc|ccc}
\toprule
\textbf{Human} 
& \textbf{I1} 
& \textbf{I2} 
& \textbf{S1} 
& \textbf{S2} 
& \textbf{Acc} 
& \textbf{F1$_{\text{avg}}$} 
& \textbf{Bal Acc} \\
\midrule
H1 & 84.00 & 74.00 & 86.67 & 29.00 & 66.39 & 64.78 & 64.81 \\
H2 & 80.00 & 75.00 & 65.00 & 30.00 & 62.22 & 59.94 & 60.31 \\
H3 & 84.00 & 88.00 & 83.33 & 24.00 & 68.33 & 65.80 & 66.13 \\

\midrule
\textbf{Average} & \textbf{82.67} & \textbf{79.00} & \textbf{78.33} & \textbf{27.67} & \textbf{65.65} & \textbf{63.51} & \textbf{63.75} \\

\bottomrule
\end{tabular}%
}
\caption{Human evaluation with performance metrics (in \%).}
\label{tab:human_eval}
\vspace{-30pt}
\end{table}
To contextualize model performance, we conduct a human evaluation on a randomly selected subset of 360 samples from test subset of the Friends dataset (100 samples from all categories, except for S1 which only had 60). 
We selected Friends for human evaluation as its social dialogue is most accessible to non-expert annotators.
Three annotators H1, H2, and H3 independently labeled each sample as \textsc{Speak} or \textsc{Silent} given the same conversation context and target speaker information provided to the models. Table~\ref{tab:human_eval} summarizes per-annotator results. Human annotators achieve 60–66\% balanced accuracy, with strong performance on I1 (explicit address) and S1 (no reference) but notably low accuracy on S2 (referenced but not addressed), the category requiring the finest pragmatic distinction. Inter-annotator agreement was moderate with the pairwise agreement scores measured by Cohen's \(\kappa\)~\cite{cohen1960agreement}, as $\kappa$ (H1-H2) = 0.57, $\kappa$ (H1-H3) = 0.38, and $\kappa$ (H2-H3) = 0.53, yielding (Avg Cohen's \(\kappa\) = 0.492), reflecting the inherent subjectivity of turn-taking decisions in multi-party settings. These results establish that this task is particularly challenging; even humans disagree on whether to speak in ambiguous situations. Notably, our best trained models (Table~\ref{tab:ft_comparison}) match or exceed human-level balanced accuracy.

\begin{table}[b!]
\vspace{-12pt}
\centering
\small
\setlength{\tabcolsep}{3pt}
\begin{tabular}{l l c c c}
\toprule
\multicolumn{5}{c}{\textbf{Training Mode Comparison}} \\
\midrule
\textbf{Dataset} & \textbf{Mode} 
& \textbf{Acc} 
& \textbf{F1$_{\text{avg}}$} 
& \textbf{Bal Acc} \\
\midrule
Friends & Decision-only & 63.64 & 63.63 & 66.60 \\
Friends & Reasoning with Decision & \textbf{70.84} & \textbf{68.80} & \textbf{68.46} \\
\midrule
\multicolumn{5}{c}{\textbf{LoRA Rank Comparison (Reasoning Mode)}} \\
\midrule
\textbf{Dataset}
&  \textbf{Rank/Alpha} & \textbf{Acc} 
& \textbf{F1$_{\text{avg}}$} 
& \textbf{Bal Acc} \\
\midrule
Friends & r=16, $\alpha$ = 32  & 67.74 & 65.47 & 65.23 \\
Friends & r=32, $\alpha$ = 64 & \textbf{70.84} & \textbf{68.80} & \textbf{68.46} \\
Friends & r=64, $\alpha$ = 128 & 69.96 & 68.29 & 68.09 \\
\end{tabular}

\vspace{4pt}

\setlength{\tabcolsep}{1.5pt}
\small
\begin{tabular}{l c c c c c c c}
\toprule
\multicolumn{8}{c}{\textbf{Combined-Dataset Training, Decision-only Mode}} \\
\midrule
\textbf{Test Dataset}
& \textbf{I1} & \textbf{I2} & \textbf{S1} & \textbf{S2}
& \textbf{Acc} 
& \textbf{F1$_{\text{avg}}$} 
& \textbf{Bal Acc} \\
\midrule
AMI     & 67.58 & 82.91 & 85.78 & 47.77 & 73.53 & 73.53 & 73.56 \\
Friends & 43.02 & 93.20 & 73.33 & 86.90 & 74.17 & 71.92 & 71.37 \\
SPGI    & 63.84 & 81.88 & 78.15 & 63.46 & 70.24 & 70.24 & 70.26 \\
\midrule
Average    & \textbf{58.15} & \textbf{86.00} & \textbf{79.09} & \textbf{66.04} & \textbf{72.65} & \textbf{71.90} & \textbf{71.73} \\
\bottomrule
\end{tabular}

\caption{Ablation analysis: training mode, LoRA rank, and combined-dataset generalization. All values are in \(\%\).}
\label{tab:mode_rank_singlecol}
\vspace{-30pt}
\end{table}

\textbf{Experiment-3 Ablation Study:}
We ablate three design choices using Qwen2.5-7B, the most stable open-source model after SFT. Table~\ref{tab:mode_rank_singlecol} (top) compares Decision-only training against the Reasoning with Decision mode on the Friends dataset. Adding reasoning traces improves accuracy by 7.2 percentage points (63.64\% to 70.84\%) and F1$_{\text{avg}}$ by 5.2 points, confirming that generating an explicit justification before the decision helps the model.
Table~\ref{tab:mode_rank_singlecol} (middle) varies the LoRA rank for the Reasoning with Decision mode on Friends dataset. Rank 32 (\(\alpha\)=64) achieves the best performance across all three metrics. Rank 16 underperforms by approximately 3 points, likely due to insufficient capacity to capture the reasoning patterns. Rank 64 shows no further gain, suggesting diminishing returns beyond rank 32 for this task.
Finally, we train a single model on the merged training splits of all three datasets to test cross-domain generalization. Table~\ref{tab:mode_rank_singlecol} (bottom) shows that pooled training achieves 71.73\% average balanced accuracy without per-domain adaptation, competitive with dataset-specific fine-tuning (Table~\ref{tab:ft_comparison}). This suggests the learned turn-taking representations transfer across conversational settings.

\section{Conclusion \& Future Work}
We formulated context-aware turn-taking as a binary prediction task for multi-party settings and introduced a 120K-sample benchmark spanning three domains. All evaluated LLMs fail under zero-shot prompting; supervised fine-tuning with reasoning distillation improves balanced accuracy by up to 23 points. Future work will incorporate multimodal cues and cross-domain generalization for real-time deployment.

\section{Acknowledgments}
We thank Busi Reddy Karnati for his generous support and contributions to the infrastructure supporting this work, and for enabling the data annotation used in this study.



\bibliographystyle{IEEEtran}
\bibliography{mybib}

\clearpage

\appendix
\begin{center}
\Large\textbf{Appendix}
\end{center}

\section{Dataset Examples}
\label{app:dataset}
In this section we show examples that demonstrate all four 
decision categories defined in Section~\ref{sec:data_creation}: 
explicit address (I1), contextual intervention (I2), no reference 
(S1), and referenced but not addressed (S2).

\begin{tcolorbox}[
    width=\linewidth,
    colback=gray!5,
    colframe=gray!60,
    boxrule=0.5pt,
    arc=2pt,
    title=\textbf{Example of Explicit Address (I1): The target
speaker is directly invited to respond.}
]

\textbf{Context:}

\textit{Chandler:} Uh, if I were omnipotent for a day, I'd..

\textit{Rachel:} See, there's always one guy. (Mocking)

\vspace{2pt}

\textbf{Current Turn:}

\textit{Monica:} Hey, Joey, what would you do if you were

\vspace{2pt}

\textbf{Target Speaker:} Joey

\vspace{2pt}

\textbf{Decision:} \textsc{SPEAK}

\vspace{2pt}

\textbf{Reasoning:} Monica directly addresses Joey and asks a question,
inviting his response to the hypothetical scenario.

\end{tcolorbox}

\begin{tcolorbox}[
    width=\linewidth,
    colback=gray!5,
    colframe=gray!60,
    boxrule=0.5pt,
    arc=2pt,
    title=\textbf{Example of Contextual Intervention (I2): The target speaker continues the interaction without being directly addressed.}
]

\textbf{Context:}

\textit{Joey:} A date?! No, no Pheebs you-you must be mistaken, because I know you wouldn't schedule a date on the same night you have plans with a friend!

\textit{Phoebe:} Come on Joey, don't make me feel badly about this.

\textit{Joey:} No, I'm gonna!! That's right! Yeah, you made me feel really guilty about goin' out with that girl!

\textit{Phoebe:} This is different! This is with David! Remember David, the scientist guy?

\textit{Joey:} Okay, well my girl from the other night was special. She was a scientist too!

\vspace{2pt}

\textbf{Current Turn:}

\textit{Phoebe:} Okay, whatever. I don't have time to convince you because he's only here for four hours, and I'm going to see him!

\vspace{2pt}

\textbf{Target Speaker:} Joey

\vspace{2pt}

\textbf{Decision:} \textsc{SPEAK}

\vspace{2pt}

\textbf{Reasoning:} Joey is actively engaged in the argument with Phoebe. 
Her final statement dismisses the unresolved conflict and prompts Joey 
to respond even though he was not explicitly addressed.

\end{tcolorbox}
\vspace{2pt}
\begin{tcolorbox}[
    width=\linewidth,
    colback=gray!5,
    colframe=gray!60,
    boxrule=0.5pt,
    arc=2pt,
    title=\textbf{Example of No Reference (S1): The target speaker is not addressed and remains silent.}
]

\textbf{Context:}

\textit{Phoebe:} Okay, cancel backup! Cancel backup!

\textit{Rachel:} Ross, didn't you say that there was an elevator in here?

\textit{Ross:} Uhh, yes I did but there isn't. Okay, here we go.

\textit{Ross:} Okay, go left. Left! Left!

\textit{Rachel:} Okay, you know what? There is no more left!

\vspace{2pt}

\textbf{Current Turn:}

\textit{Rachel:} Any chance you think the couch looks good there?

\vspace{2pt}

\textbf{Target Speaker:} Phoebe

\vspace{2pt}

\textbf{Decision:} \textsc{SILENT}

\vspace{2pt}

\textbf{Reasoning:} Phoebe is a bystander in this exchange. Rachel’s
question is directed toward Ross, who is actively moving the couch with
her, and Phoebe has not been involved in the recent interaction.

\end{tcolorbox}
\vspace{5pt}
\begin{tcolorbox}[
    width=\linewidth,
    colback=gray!5,
    colframe=gray!60,
    boxrule=0.5pt,
    arc=2pt,
    title=\textbf{Example of Referenced but not addressed (S2): The target speaker is part of the interaction but does not take the next turn.}
]

\textbf{Context:}

\textit{Chandler:} No, no, no, no, no, NO! No, no... we're not together. We're not a couple. We're definitely not a couple.

\textit{Joey:} Well, you seem pretty insulted by that. What? I'm not good enough for you?

\textit{Chandler:} We're not gonna have this conversation again... Look at this place. Why am I so intimidated by this guy? Pretentious art, this huge macho couch. When we know all he does is sit around all day crying about losing Monica to a real man! (laughs) You don't think he's here, do you?

\textit{Joey:} You know what it is? It's a nice place but I gotta say I don't know if I see myself living here. Oh, oh, oh, let me see... (Joey sits down on the couch, mimes opening a can and puts his hand down his pants) Yeah, I could see it.

\textit{Chandler:} Look at these videos. You know, I mean, who does he think he is? Magnum Force, Dirty Harry, Cool Hand Luke... Oh my God!

\textit{Chandler:} There's a tape here with Monica's name on it.

\textit{Joey:} Ooh! A tape with a girl's name on it. It's probably a sex tape... Wait a minute... This says Monica... And this is Richard's apartment...

\vspace{2pt}

\textbf{Current Turn:}

\textit{Chandler:} Get there faster Joey! (Joey gasps and finally understands.)

\vspace{2pt}

\textbf{Target Speaker:} Joey

\vspace{2pt}

\textbf{Decision:} \textsc{SILENT}

\vspace{2pt}

\textbf{Reasoning:} Joey is an active participant in the conversation and Chandler’s remark is directed at him to prompt realization about the situation. Joey reacts non-verbally (gasping), indicating understanding, but he does not take the next spoken turn, allowing Chandler to continue speaking.

\end{tcolorbox}

\onecolumn
\section{Prompts}
\label{app:prompts}

\begin{tcolorbox}[
    width=\textwidth,
    title=Instruction Prompt,
    colback=gray!5,
    colframe=gray!60,
    boxrule=0.5pt,
    arc=2pt
]
\ttfamily\footnotesize
You are playing the role of Speaker \{target\_speaker\}. The conversation history above shows all utterances including the most recent one (marked as [MOST RECENT]). After that most recent utterance, there was a pause. Decide if you (Speaker \{target\_speaker\}) should START TALKING or STAY SILENT now.
\end{tcolorbox}

\vspace{4pt}
\begin{tcolorbox}[
    width=\textwidth,
    title=System Prompt,
    colback=gray!5,
    colframe=gray!60,
    boxrule=0.5pt,
    arc=2pt
]

\ttfamily\footnotesize

You are a helpful, concise assistant.
You are a turn-taking decision model in a multi-party conversation where multiple people are talking.
You are roleplaying the role of the target speaker you are given.
Your job is to decide whether the target speaker should START TALKING or STAY SILENT after a detected pause in conversation.

You will receive:
\begin{enumerate}
\item An instruction telling you the target speaker role (e.g., "Speaker C", "Speaker X", or "Nova")
\item The previous conversation context with speaker-labeled transcript
\item Most recent utterance: the most recent utterance said in the conversation, after which you have to make a decision
\end{enumerate}

First, determine the target speaker's ROLE in the current exchange:\\
- ACTIVE PARTICIPANT: The target speaker has been speaking, was addressed, or is part of an ongoing back-and-forth in the current topic.\\
- BYSTANDER: The target speaker has not been involved in the current exchange and is passively listening.\\

RULES FOR DECIDING:

Output SILENT when:\\
- The target speaker is a BYSTANDER and the recent utterance is directed at someone else\\
- The target speaker has not been referenced, addressed, or involved\\
- The recent utterance is clearly incomplete\\
- Someone mentions the target speaker in third person without expecting a response\\

Output SPEAK when:\\
- The recent utterance directly addresses the target speaker with a question or request\\
- The recent utterance asked the target speaker something and this is a clear follow-up\\
- The context makes it unambiguous the speaker is waiting for the target speaker\\
- The speaker redirects the conversation to the target speaker\\
- The recent utterance is a group-directed question and the target speaker is part of the group\\
- The target speaker is an ACTIVE PARTICIPANT and the utterance completes a thought requiring response\\
- The target speaker previously asked a question and the recent utterance answers it\\
- Staying silent would unnaturally drop them from the conversation\\
- A natural backchannel or reaction is expected\\

IMPORTANT NUANCES:\\
- The key distinction is ACTIVE PARTICIPANT vs BYSTANDER.\\
- When uncertain and the target speaker is a BYSTANDER → prefer SILENT.\\
- When uncertain and the target speaker is ACTIVE → check if the utterance is directed elsewhere.\\
- False interruptions are bad, but failing to respond when involved also breaks the interaction.\\

Output format:

<reasoning>One sentence explaining whether the target speaker is an ACTIVE PARTICIPANT or BYSTANDER and whether they should respond.</reasoning>  \\
<decision>SPEAK</decision> or <decision>SILENT</decision>  \\
<confidence>high, medium, or low</confidence> \\

CRITICAL: The decision tag must contain only SPEAK or SILENT.

EXAMPLES:

Example 1  
Target speaker: Alex  \\
Speakers: Alex, Sam  \\
Recent utterance (Sam): "Wait, you actually told her?"  

<reasoning>Alex is an ACTIVE PARTICIPANT and Sam directly asks Alex a question.</reasoning>  \\
<decision>SPEAK</decision>  \\
<confidence>high</confidence> \\

Example 2  
Target speaker: Jordan  \\
Speakers: Alex, Jordan, Sam  \\
Recent utterance (Alex): "Yeah, it was rough. I didn't sleep at all last night."  

<reasoning>Jordan is a BYSTANDER and Alex is narrating to the group.</reasoning>  \\
<decision>SILENT</decision>  \\
<confidence>high</confidence> \\

\end{tcolorbox}

\vspace{-0.3cm}


\end{document}